\documentclass[runningheads]{llncs}
\usepackage{graphicx}

\usepackage{tikz}
\usepackage{comment}
\usepackage{amsmath,amssymb} 
\usepackage{color}
\usepackage{amsmath,graphicx}
\usepackage{booktabs}
\usepackage{multirow}
\usepackage{caption}
\usepackage{subcaption}
\usepackage{amsmath}
\usepackage{amssymb}
\usepackage{comment}
\usepackage{float}
\usepackage{booktabs}
\usepackage{cite}

\begin{document}
\setcounter{secnumdepth}{3}

\pagestyle{headings}
\mainmatter
\def\ECCVSubNumber{100}  

\title{Privacy-Preserving In-Bed Pose Monitoring: \\ A Fusion and Reconstruction Study} 

\titlerunning{Privacy-Preserving In-Bed Pose Monitoring}

\author{Thisun Dayarathna\inst{1} \and
Thamidu Muthukumarana\inst{1} \and
Yasiru Rathnayaka\inst{1} \and
Simon Denman\inst{2} \and
Chathura de Silva\inst{1} \and
Akila Pemasiri\inst{1,2} \and
David Ahmedt-Aristizabal\inst{2,3}
}
\authorrunning{Dayarathna et al.}
%
\institute{Universtiy of Moratuwa, Moratuwa, Sri Lanka 
\email{\{thisun.17,thamidurm.17,yasirurathnayaka.17,akila.10\}@cse.mrt.ac.lk}
\and
Queensland University of Technology, Brisbane, Australia \\
\email{s.denman@qut.edu.au} 
\and
Imaging and Computer Vision group, CSIRO Data61, Canberra, Australia\\
\email{david.ahmedtaristizabal@data61.csiro.au} 
}
\maketitle

\begin{abstract}
\vspace{-2pt}
Recently, in-bed human pose estimation has attracted the interest of researchers due to its relevance to a wide range of healthcare applications. Compared to the general problem of human pose estimation, in-bed pose estimation has several inherent challenges, the most prominent being frequent and severe occlusions caused by bedding.
In this paper we explore the effective use of images from multiple non-visual and privacy-preserving modalities such as depth, long-wave infrared (LWIR) and pressure maps for the task of in-bed pose estimation in two settings. First, we explore the effective fusion of information from different imaging modalities for better pose estimation. Secondly, we propose a framework that can estimate in-bed pose estimation when visible images are unavailable, and demonstrate the applicability of fusion methods to scenarios where only LWIR images are available. 
We analyze and demonstrate the effect of fusing features from multiple modalities. For this purpose, we consider four different techniques: 1) Addition, 2) Concatenation, 3) Fusion via learned modal weights, and 4) End-to-end fully trainable approach; with a state-of-the-art pose estimation model.
We also evaluate the effect of reconstructing a data-rich modality (\textit{i.e.}, visible modality) from a privacy-preserving modality with data scarcity (\textit{i.e.}, long-wavelength infrared) for in-bed human pose estimation. For reconstruction, we use a conditional generative adversarial network.
We conduct experiments on a publicly available dataset for feature fusion and visible image reconstruction. We conduct ablative studies across different design decisions of our framework. This includes selecting features with different levels of granularity, using different fusion techniques, and varying model parameters. Through extensive evaluations, we demonstrate that our method produces on par or better results compared to the state-of-the-art.
The insights from this research offer stepping stones towards robust automated privacy-preserving systems that utilize multimodal feature fusion to support the assessment and diagnosis of medical conditions.

\vspace{-5pt}
\end{abstract}

\section{Introduction}
\label{sec:intro}

In-bed human pose estimation has demonstrated its utility for many healthcare applications, including remote patient monitoring and preventing conditions such as carpal tunnel syndrome, pressure ulcers, and sleep apnea~\cite{mccabe2007epidemiologic}. 

Skeleton-based top-down models are widely used in recent pose estimation works due to their simple and flexible representation~\cite{chen2020monocular}.
Many recent pose estimation network designs~\cite{8953615, xiao2018simple, cao2017realtime, newell2016stacked} have achieved improved performance by introducing multi-scale feature processing mechanisms, where features at different scales are utilized in parallel.
Among these approaches, high-resolution net (HRNet)~\cite{8953615} has achieved significant popularity over the past two years due to its fast training and inference speed, and its ability to maintain a high-resolution feature representation throughout the network, leading to improved localisation accuracy.

Despite many works reporting accurate estimation on publicly available general human pose datasets, in-bed human pose estimation tasks come with unique challenges due to the environment setting and the limited availability of datasets with in-bed human poses. The most prominent challenges are:
1) occlusion due to the human body being covered by bedding, and 
2) capturing human in-bed poses using visible modalities within a hospital environment which is constrained by poor lighting conditions and where privacy concerns are present. 

Most existing methods for in-bed pose estimation are driven by deep learning models to address these difficulties, and seek to close the gap between general data and in-bed human pose data~\cite{8265373,8490852,liu2019bed,ahmedt2018deep,ahmedt2019understanding,wang2021video}. Such methods can be categorized into two main domains: fine-tuning and feature fusion.

The majority of literature focuses on using a single modality for in-bed pose estimation and relies on fine-tuning. Liu \textit{et al.}~\cite{8265373} introduced a vision based hierarchical inference model for in-bed pose classification using video sequences from the visible modality. Histogram of Oriented Gradients (HOG) features are extracted from each frame and fed to a pre-trained model for posture classification. 
Chen \textit{et al.}~\cite{8490852} introduced an extension to an existing Convolutional Neural Network (CNNs) pose-based framework~\cite{pfister2015flowing} for accurate upper body posture estimation in a clinical environment, using video sequences captured in the visible modality. 
Liu \textit{et al.}~\cite{liu2019bed} adopted the Convolutional Pose Machine~\cite{wei2016convolutional} for in-bed pose estimation by fine-tuning the model with their Infra-Red Selective (IRS) dataset.
Such fine-tuning approaches for in-bed human pose estimation have been used in different healthcare applications such as seizure disorder classification~\cite{ahmedt2018deep,ahmedt2019understanding} and inpatient fall risk assessment~\cite{wang2021video}.

Multi-modal data introduces redundancy and allows methods to compensate for errors in some modalities, leading to better diagnostic decisions \cite{pemasiri2021multi, pemasiri2019semantic}. For instance, lighting conditions may impede visible images, but will not affect pressure images. In addition, since the availability of different sensors and patient preferences may vary, scenarios in which unimodal pose estimation methods can be applied may be limited.

When considered alone, any given modality has desirable and undesirable properties. For instance, although Long Wave Infrared (LWIR) can see through covers, it may be affected by residual heat of the body, or heat from other objects. While visible images are not affected by residual heat, it is more susceptible to pose estimation errors when joints are covered (\textit{i.e.} by bedding) compared to the LWIR modality \cite{yin2020multimodal}. Similarly, while the pressure modality does not detect all joints in all body configurations, pressure sensor signals typically do not change with cover conditions. The use of multi-modal data allows the deficiencies of individual modes to be addressed by each other. Furthermore, capturing visible images of patients in a hospital environment is generally less acceptable than other modalities, as visible images lead to privacy concerns as the subject's identity is exposed. 

Despite the promise of multi-modal methods, existing works that effectively utilize complementary information resulting from the fusion of different modalities for in-bed pose estimation are less common than their uni-modal counterparts. 
Karanam \textit{et al.}~\cite{karanam2020towards} and Yin \textit{et al.} \cite{yin2020multimodal} used multi-modal data for patient positioning and shape estimation. Karanam \textit{et al.} introduced a separate feature fusion module to extract features from different modalities and generate  fused feature vectors. In contrast, in our approach we extract features from intermediate stages of uni-modal HRNet models without using a new module. In comparison to Yin \textit{et al.}, where feature extraction is done at subsequent stages, our approach applies fusion to a specific stage of the HRNet model. 

In a preliminary work, we have addressed human pose analysis on privacy-preserving data from an unsupervised perspective. To this end, a novel multimodal conditional variational autoencoder (MC-VAE) was introduced for human pose estimation ~\cite{cao2021bed}. The MC-VAE is capable of reconstructing features of missing modalities using the available features. Hence, this method can be used for accurate pose estimation when only a privacy-preserving modality (LWIR) is available, while leveraging the reconstructed features of the visible modality. During the training phase, the feature distribution of the visible modality is learned by the feature reconstruction module. During the inference, visible features are rebuilt conditioned on the LWIR modality.
In this work, we propose using a conditional Generative Adversarial Network (cGAN) to reconstruct non-privacy preserving visible images using privacy preserving LWIR images. Here, a generator is trained to learn the mapping between visible images and LWIR images, and during the inference stage, the generator generates synthetic visible images based on the learned feature mapping, and the generated images are fused with original LWIR images for pose estimation. As such, the fusion of multiple modes can still be used to improve performance, even when only a single mode is available.

In this work, we exploit the discerning features captured by each modality and combine them through feature fusion such that multi-modal data can be effectively used for in-bed human pose estimation on a publicly available dataset. This case study focuses on feature fusion of two modalities, as this is more readily adaptable to low resource clinical environments. However, our proposed methods can be extended to multiple synchronized modalities.
We also introduce fusion approaches for pose estimation using visible images reconstructed from LWIR images. Translating LWIR images to the visible domain generates synthetic images that obscure the facial identity of the patient, thus addressing privacy issues, yet enabling the proposed fusion approach when visible images are unavailable. We focus only on visible image reconstruction from LWIR in this study, as visible images pose greater privacy concerns. We select LWIR images as the reconstruction source in this study as they can be captured at a low cost compared to other modalities, and they can mitigate bedding occlusions.

Our main contributions are summarized as follows:
\begin{enumerate}
\vspace{-2pt}
\item We propose intermediate fusion techniques for in-bed pose estimation, that integrate high-level features from multiple modalities, and investigate the effect of the collective manipulation of the diverse information available from each modality.

\item We demonstrate our feature fusion methods' ability to generalize by applying it to two portions of the simultaneously-collected multi-modal Lying Pose (SLP) dataset \cite{liu20120simultaneously, liu2019seeing} with adult subjects, covering two different environment settings, defined as ``home" and ``hospital".

\item We exploit conditional generative adversarial networks (cGAN) to generate synthetic visible images from LWIR images (\textit{i.e.,} LWIR to visible image reconstruction), which encourages privacy preservation. 
We explore the applicability of these synthetic visible images for pose estimation using Visible-LWIR fusion, when visible images are not available.
\end{enumerate}

The remainder of the paper is organized as follows. Section \ref{methodology} presents the methodology for the fusion techniques and synthetic visible image generation for pose prediction. Section \ref{experiments} presents our experiments including details of the datasets used, the evaluation protocol, and qualitative and quantitative results. Finally, Section \ref{conclusion} concludes the paper with a discussion of the presented methodologies and potential future directions for this research.

\section{Methods}
\label{methodology}

In this section, we describe the methodology for feature fusion and visible image reconstruction from LWIR images. 
In Subsection \ref{network_architecture}, we describe the backbone architecture used in this paper, Subsection \ref{fusion_appropach} discusses different feature fusion techniques that we adopted in our work. Subsection \ref{visible_reconstruction_from_LWIR} outlines the reconstruction of visible images from LWIR images.

\begin{figure*}[!t]
\begin{center}
\includegraphics[width=1\linewidth]{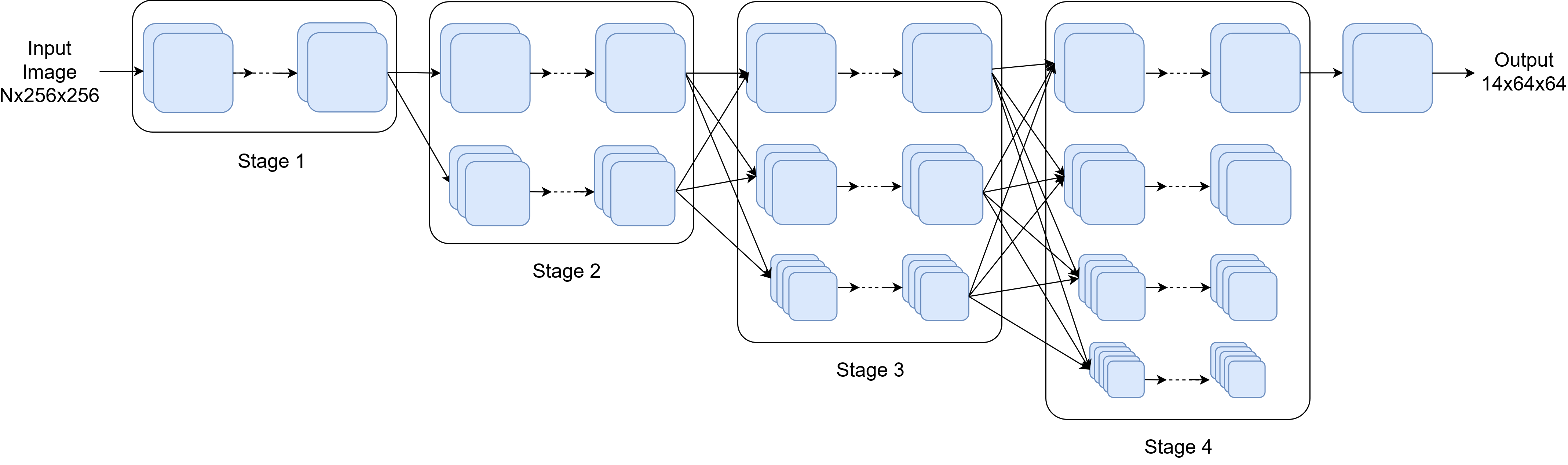}
\end{center}
\vspace{-15pt}
  \caption{The HRNet architecture consists of 4 stages. In each stage, branches are arranged to process high-to-low resolution feature maps. At the end of each stage except the final stage, the information is exchanged between branches of the next stage. The input to the model is an $N$ $\times$ 256 $\times$ 256 image, where $N$ is the number of image channels and varies with the modality (for the visible modality $N$=3, for other modalities $N$=1). The generated output contains 14 64 $\times$ 64 heatmaps, corresponding to 14 predicted key points. Recreated from~\cite{8953615}.}
\label{HRNet}
\vspace{-10pt}
\end{figure*}

\vspace{-5pt}
\subsection{Base network architecture}
\label{network_architecture}

We adopt HRNet \cite{8953615} as the backbone network architecture for human pose estimation as it is currently the SOTA model for pose estimation in terms of efficiency and accuracy. The architecture is illustrated in Fig.~\ref{HRNet}. 
HRNet uses a top-down method, and uses both high and low resolution images to enhance position sensitivity and recognition capability. The network consists of 4 stages and 4 parallel sub-networks (branches), where the resolution is halved, and the number of channels is doubled in each subsequent branch. The input to the network is an $N$ $\times$ 256 $\times$ 256 image, where $N$ is the number of channels in the image and varies with the modality. The output is 14 heatmaps, each of size 64 $\times$ 64, corresponding to 14 joints on the human body.

We evaluate the fusion of features extracted from multiple internal stages of the HRNet model to examine the effect of the granularity of the extracted features on fusion performance.
Hereafter, in this paper ``stage'' refers to the stages in the HRNet model as shown by Fig.~\ref{HRNet}.

\vspace{-5pt}
\subsection{Fusion approaches}
\label{fusion_appropach}

Usually, in multi-modal feature fusion, the same scene captured by different modalities is provided as the input, and the feature representations of the modes are extracted via deep learning models. Then, these generated features are combined to obtain improved features. This method helps to learn features which are more discriminative. In our method, feature fusion is used to merge independent features extracted from different modalities into a single unique feature set, which can then be used within the remainder of a single mode's pipeline.

As mentioned above, few related works use feature fusion for in-bed pose estimation. 
Karanam \textit{et al.}\cite{karanam2020towards} proposed an intermediate fusion technique where a feature fusion module generates a new feature representation for the joint feature space extracted from multiple modalities. 
In our proposed method, we extract features from intermediate stages of the uni-modal HRNet models. These extracted features are then fused and fed into the appropriate next stage of the same baseline model without introducing any new modules. 

Yin \textit{et al.} \cite{yin2020multimodal} proposed a pyramid fusion scheme, where modalities are fused sequentially at different levels for shape estimation. The authors initially concatenated the most informative modalities (depth and LWIR) in the SLP dataset, to get a rough shape estimation. Then, the rough estimation is sequentially refined by feeding it to another model stage. The input to each subsequent stage is the current rough shape estimation and the concatenation of the previous stage's input with a new modality.
Unlike \cite{yin2020multimodal}, our method fuses the features extracted from different modalities within a single network instead of using multiple subsequent refining stages. In addition, \cite{yin2020multimodal} used early fusion, relying on concatenation at the beginning of each stage of the approach.

In contrast to these approaches, we reap the benefits of one of the best performing architectures for general pose estimation (\textit{i.e.}, HRNet \cite{8953615}), and exploit intermediate fusion by fusing features extracted from pre-trained HRNet models at either stage 2 or stage 3.
For our feature fusion experiments, we first selected a set of modalities, $M$, for fusion. Then, we applied our multi-modal fusion method to the output of a single stage. When fusing the outputs of the $N^{th}$ stage, we use $|M|$ (number of modalities) pre-trained uni-modal HRNet models, which comprise only the first $N$ stages. These uni-modal models are not retrained in our first three fusion methods. The outputs from the first $N$ stages of these uni-modal models are fused to obtain a fused $N^{th}$ stage output, based on different fusion techniques, discussed later in this section. We experimented with $N = 2$ and $N = 3$. Therefore, we fuse feature tensors produced by the two branches of stage two (\( O_{2,1}^{modality}\), \( O_{2,2}^{modality}\)) or the three branches of stage three (\( O_{3,1}^{modality}\), \( O_{3,2}^{modality}\), \( O_{3,3}^{modality}\)) of the pre-trained HRNet models for the considered modalities. 
As an example, all the three branches in stage three of the depth modality (\( O_{3,1}^{depth}\), \( O_{3,2}^{depth}\), \(O_{3,3}^{depth}\)) are fused with each of the stage three branches of the LWIR modality (\( O_{3,1}^{LWIR}\), \( O_{3,2}^{LWIR}\), \( O_{3,3}^{LWIR}\)).
Then this fused $N^{th}$ stage output is fed to HRNet stage $N+1$ (in this case, stage four). Only the weights of the layers following stage $N$ will be updated during training.

We use addition and concatenation as the basis for our feature fusion, as fusion using element-wise addition of features \cite{7478072,10.1007/978-3-319-46493-0_38,lin2017feature, deep_residual} and concatenation of features \cite{7298594,8099726} has been successfully utilized in many deep learning applications. These approaches are employed as follows.

\subsubsection{Addition:} \label{add_desc} 
\begin{figure*}[!t]
\begin{center}
\includegraphics[width=0.95\linewidth]{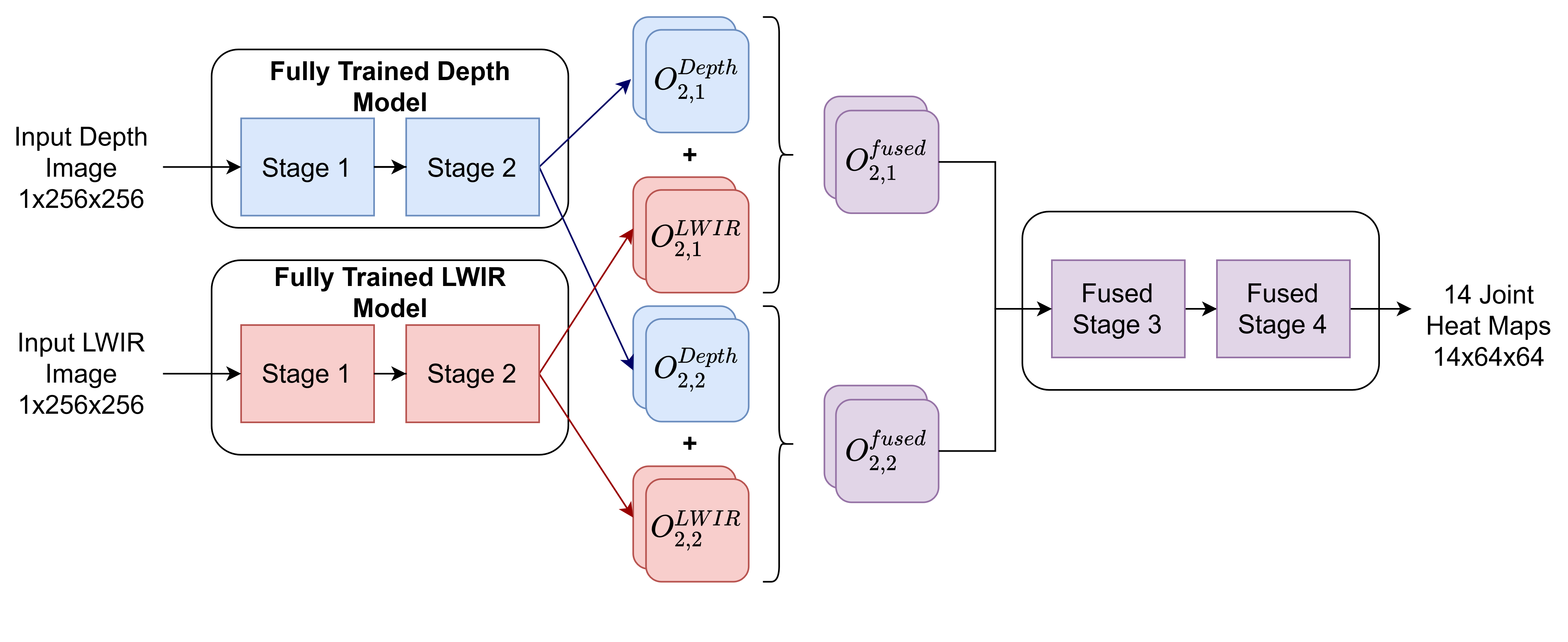}
\end{center}
\vspace{-23pt}
  \caption{Intermediate fusion using addition at stage two of HR-NET with LWIR and depth modalities.}
\label{add}
\vspace{-10pt}
\end{figure*}

In this approach, extracted feature vectors from different modalities are subjected to the addition operation. Then, the resulting fused feature vectors at stage $N$ are directly fed into the $N+1$th stage of the HRNet model. An example of feature fusion based on the addition from stage two features is depicted in Fig.~\ref{add}.

The fusion output ($O_{N,b}^{fused}$) that results from fusing the $b^{th}$ branches in the $N^{th}$ stage of all modalities can be represented as follows,
\begin{equation}
\label{unweighted_add}
    O_{N,b}^{fused} = \sum_{m {\;}\in {\;}M} O_{N,b}^m . 
\end{equation}

In this summation, the addition of two branch outputs (\(O_{N,b}^{depth} + O_{N,b}^{LWIR}\)) represents the element-wise addition of each corresponding channel pair of the two branch outputs. The resultant set of channels is the fused output. Similarly, this operation can be extended to $|M| > 2$.

\subsubsection{Concatenation:} \label{concat_desc} 
\begin{figure*}[!t]
\begin{center}
\includegraphics[width=0.95\linewidth]{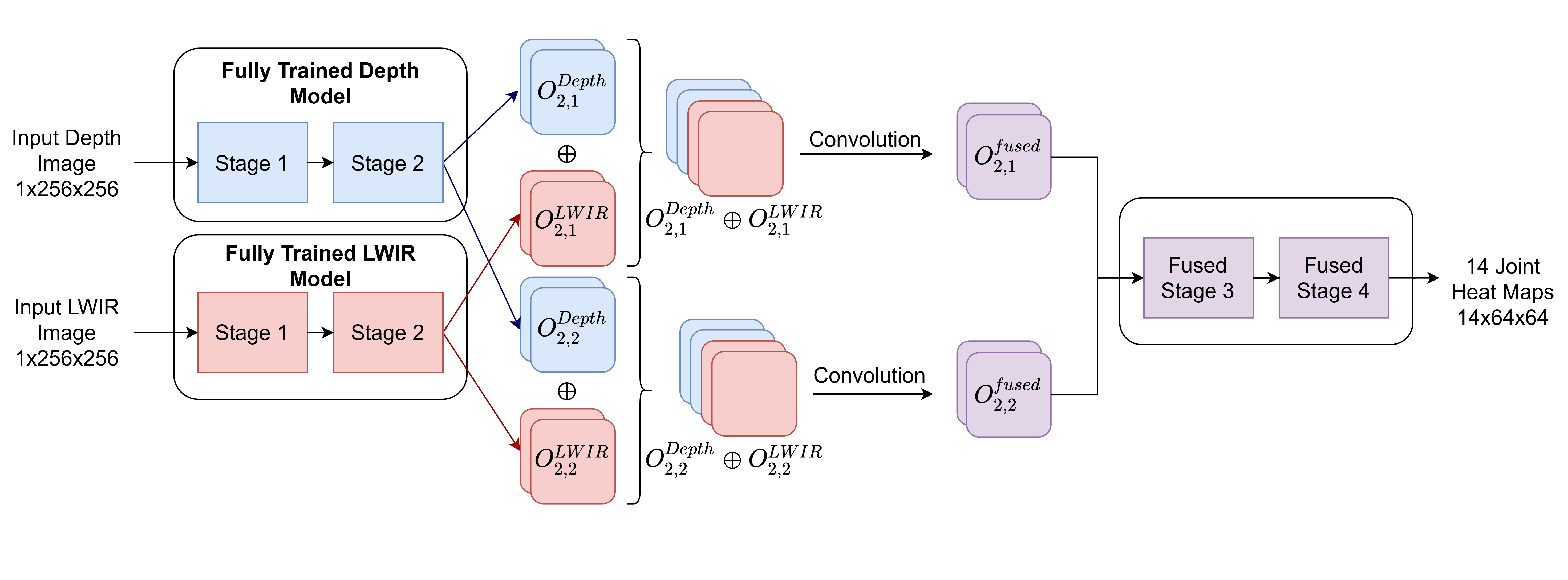}
\end{center}
\vspace{-23pt}
  \caption{Intermediate fusion using concatenation at stage two of HR-NET with LWIR and depth modalities.}
\label{concat}
\vspace{-10pt}
\end{figure*}

In this approach, extracted feature vectors from different modalities are subjected to the concatenation operation.
An additional convolutional operation is added to reduce the number of channels to that expected by stage $N+1$ of HRNet. An example of feature fusion based on the concatenation of stage two features is depicted in Fig.~\ref{concat}.

The fusion output ($O_{N,b}^{fused}$) from fusing the $b^{th}$ branches in the $N^{th}$ stage of all modalities can be represented as follows,
\begin{equation}
\label{unweighted_concat}
O_{N,b}^{fused} =  CNN_{N,b}(O_{N,b}^{m_1} \oplus O_{N,b}^{m_2} \oplus ... \oplus O_{N,b}^{m_n}), \\ 
where{\;}m_1, m_2, ... ,m_n {\;} \in {\;} M .
\end{equation}

The output of the \(\oplus\) operation on the two branch outputs (\(O_{N,b}^{depth} \oplus O_{N,b}^{LWIR}\)) is the tensor created by stacking the channels of the two branch outputs together. Similarly, this operation can be extended to $|M| > 2$. $CNN_{N,b}$ represents the convolutional layer with the kernel size of $1 \times 1 $ used to reduce the channel depth of the concatenated tensor to the dimensions of branch $b$ of stage $N$ of HRNet.

\subsubsection{Fusion with learned modal weights:} \label{weighted_desc} 
We introduce a set of trainable weights to exploit the different contributions offered by features extracted from each modality towards the final prediction. A learnable weight vector, $w_b^{modality}$, of size $n_b$ is introduced for each branch from each modality. Here $n_b$ is the number of channels in the corresponding branch.
 
For this approach, each extracted channel from each branch of each modality is multiplied by the corresponding weight before the fusion operation (\textit{i.e.}, addition or concatenation). 
This penalizes feature maps that negatively contribute to the final prediction but reward those that positively contribute. As such, this can be seen as a channel-wise attention process. The weight for each channel is initialized at $1$ for the feature maps extracted from the modality with the highest testing accuracy, and $0$ for the other.

\begin{figure*}[!t]
\begin{center}
\includegraphics[width=1\linewidth]{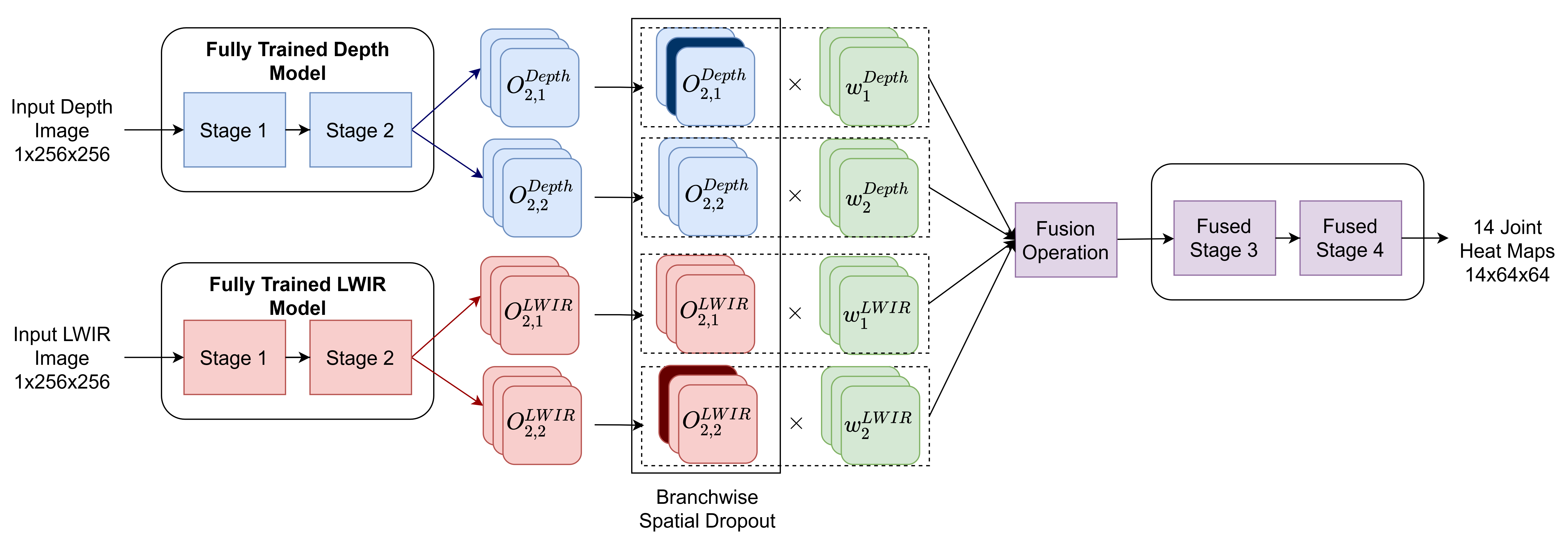}
\end{center}
\vspace{-15pt}
  \caption{Fusion with learned channel-wise mode weights, applied to stage two of HR-Net with LWIR and depth modalities.}
\label{WeightNet}
\vspace{-10pt}
\end{figure*}

To reduce overfitting, each set of channels from each branch from each modality is randomly zeroed using spatial dropout\cite{tompson2015efficient} before multiplying the branch outputs with the corresponding weight. Channels are dropped with a probability of 0.2, sampled from a Bernoulli distribution. Outputs are also batch normalized \cite{batch_norm} before feeding information to the next layer, and a ReLU is applied.
This fusion method is illustrated in Fig.~\ref{WeightNet}.
 
Equations \ref{unweighted_add} and \ref{unweighted_concat} are modified as follows to incorporate the weighting technique. The $\Omega(\cdot)$ operator represents the spatial dropout operation,
\begin{equation}
\label{weighted_add}
    O_{N,b}^{fused} = \sum_{m \in M} w_b^m \Omega(O_{N,b}^m) ,
\end{equation}
\begin{equation}
\label{weighted_concat}
\resizebox{0.85\textwidth}{!}{$O_{N,b}^{fused} = CNN_{N,b}(w_b^{m_1}\Omega(O_{N,b}^{m_1}) {\;} \oplus ... \oplus {\;} w_b^{m_n}\Omega(O_{N,b}^{m_n})), \\ 
where {\;} m_1, ... ,m_n \in M $}.
\end{equation}

\subsubsection{End-to-end fully trainable approach:} \label{end-to-end_desc} 
\begin{figure*}[!t]
\begin{center}
\includegraphics[width=1\linewidth]{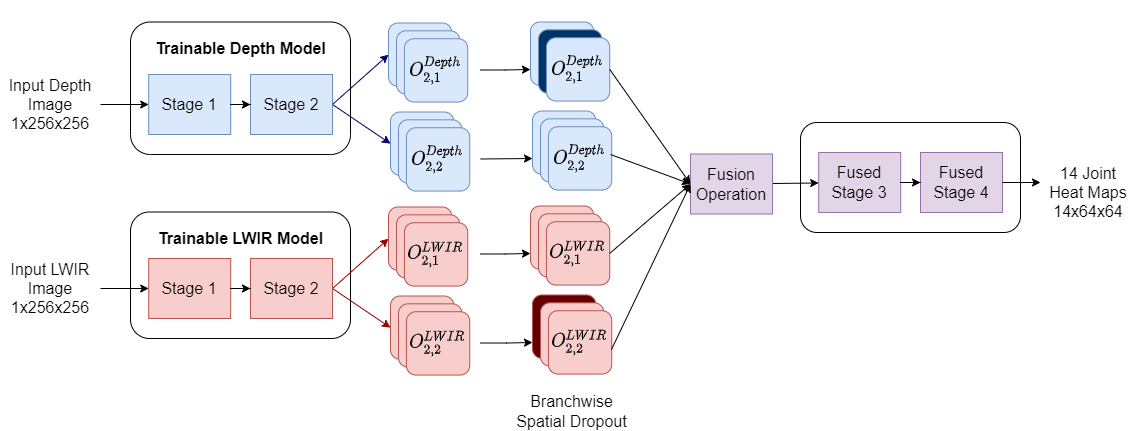}
\end{center}
\vspace{-15pt}
  \caption{End-to-end fully trainable approach applied to stage two of HR-NET with LWIR and depth modalities.}
\label{end-to-end}
\vspace{-10pt}
\end{figure*}

In the approaches described in Subsection~\ref{add_desc}, Subsection~\ref{concat_desc}, and Subsection~\ref{weighted_desc}, we used frozen models for feature extraction before fusion. But within the end-to-end approach, we allow the stages before the fusion to be trained as well. This fusion method is illustrated in Fig.~\ref{end-to-end}.
Learnable weights are not used in this method, as the trainable weights in the previous stages can model the importance of each modality implicitly. However, the spatial dropout operation, batch normalization, and ReLU operation are used similarly to the way they are used in Subsection~\ref{weighted_desc}.

Equations \ref{unweighted_add} and \ref{unweighted_concat} are modified as follows in this method. The $\Omega(\cdot)$ operator represents the spatial dropout operation,
\begin{equation}
\label{e2e_add}
    O_{N,b}^{fused} = \sum_{m \in M} \Omega(O_{N,b}^m) ,
\end{equation}
\begin{equation}
\label{e2e_concat}
\resizebox{0.85\textwidth}{!}{$O_{N,b}^{fused} = CNN_{N,b}(\Omega(O_{N,b}^{m_1}) {\;} \oplus ... \oplus {\;}\Omega(O_{N,b}^{m_n})), \\ 
where {\;} m_1, ... ,m_n \in M $}.
\end{equation}

\textit{Remarks on fusion approaches:}
Compared to the addition technique, the concatenation technique better allows the preservation of semantic information. On the other hand, adding features can help to reduce the number of parameters in the model compared to concatenation, and thus speed up training and inference. Furthermore, fusion with the learned model weights can better capture salient information from different modalities, while the end-to-end fully trainable method allows the base network to learn the features that are more specific to the fusion setting.

All fusion approaches can be extended to more than two modalities. To enable this, feature maps captured from the intermediate stages of additional uni-modal HRNet models can be included in the fusion operation as explained above. 
We demonstrate the ability of the proposed techniques to generalize in Section \ref{experiments}.

\vspace{-5pt}
\subsection{Visible image reconstruction from LWIR images}
\label{visible_reconstruction_from_LWIR}

We explore the possibility of generating privacy-preserving synthetic visible images using data from the LWIR modality to enable data fusion in a scenario where only the LWIR modality is available. To evaluate the effectiveness of this approach, we used the reconstructed visible images in fusion with the LWIR modality. 

Generative Adversarial Networks (GANs) \cite{goodfellow2014generative} have been widely adopted for many applications across many domains \cite{liang2021swinir, zhou2018stock, guo2017synthetic}. For example, Isola \textit{et al.}~\cite{pix2pix2017} proposed a variation of the conditional GAN (cGAN)~\cite{mirza2014conditional} which has been successfully adapted to many cross-domain learning tasks \cite{pemasiri2021im2mesh, huang2017arbitrary, zhu2017unpaired, wang2018high}. 

To generate synthetic visible images from LWIR images, we adapt the unconditional GAN proposed in~\cite{pix2pix2017}, referred to as ``pix2pix'' from here onward. 
The visible images generated by this approach, using the LWIR model as stimulus, obscure the patient's identity.
Furthermore, the capability of the pix2pix model to learn a generalized mapping between images of different modes makes it a suitable candidate for image reconstruction.
Our adaptation of the loss functions in~\cite{pix2pix2017} is described below.

In an unconditional GAN, the discriminator, $D$, tries to separate the real visible images, ${Im_{visible}}$, from the synthetic visible images, $G(Im_{LWIR},z)$, where the synthetic images are the output of the generator, $G$. Hence, the objective function for an unconditional GAN is as follows,
\begin{equation} 
\resizebox{0.92\textwidth}{!}{$\mathcal{L}_{GAN}\left ( G,D \right ) = \mathbb{E}_{Im_{visible}}\left [logD\left ( Im_{visible} \right )  \right ]  + \mathbb{E}_{Im_{LWIR},z}\left [log\left ( 1- D\left ( Im_{LWIR},G\left(Im_{LWIR},z \right) \right ) \right ) \right] $}.  
\label{GAN_equation}
\end{equation}

When conditioning is taken into account, as in the pix2pix model, the objective of $D$ is to separate the synthetic visible image generated by $G$ conditioned on the input LWIR image $Im_{LWIR}$, from the real visible image $Im_{visible}$ conditioned on the same input LWIR image. At the same time, the generator $G$ tries to beat the discriminator, $D$, by making the generated visible image appear to be a real visible image. This is captured by the following objective function:
\begin{equation} 
\resizebox{0.92\textwidth}{!}{$\mathcal{L}_{cGAN}\left ( G,D \right ) = \mathbb{E}_{Im_{LWIR},Im_{visible}}\left [logD\left ( Im_{LWIR},Im_{visible} \right )  \right ]  + \mathbb{E}_{Im_{LWIR},z}\left [log\left ( 1- D\left ( Im_{LWIR},G\left(Im_{LWIR},z \right) \right ) \right ) \right] $}.  
\label{conditional_equation}
\end{equation}

In the pix2pix model, $G$ needs to generate visible images similar to the ground truth and mislead $D$ about the authenticity of the generated image. This is done by trying to minimize the L1 loss between generated visible image and real visible image,
\begin{equation} 
\mathcal{L}_{L_{1}}\left ( G\right ) = \mathbb{E}_{Im_{LWIR},Im_{visible},z}\left [ \left \| Im_{visible}-G\left ( Im_{LWIR},z \right ) \right \|_{1} \right ] ,
\label{Gen_loss_equation}
\end{equation}

The final objective of the pix2pix model is thus
\begin{equation} 
G^{*} = arg\;\underset{G}{min}\;\underset{D}{max} \mathcal{L}_{cGAN}\left ( G,D \right )+\lambda \mathcal{L}_{L1}(G) .
\label{objective_function}
\end{equation}

It is important to highlight that although we generate visible images reconstructed from LWIR images, it is possible to adapt Equation \ref{conditional_equation} and Equation \ref{Gen_loss_equation} to translate between any combination of modalities. Equation \ref{conditional_equation_general} and Equation \ref{Gen_loss_equation_general} generate images of modality $j$ ($Im_{j}$) from images of modality $i$ ($Im_{i}$).
\begin{equation} 
\resizebox{0.9\textwidth}{!}{$\mathcal{L}_{cGAN}\left ( G,D \right ) = \mathbb{E}_{Im_{j},Im_{i}}\left [logD\left ( Im_{j},Im_{i} \right )  \right ]  + \mathbb{E}_{Im_{j},z}\left [log\left ( 1- D\left ( Im_{j},G\left(Im_{j},z \right) \right ) \right ) \right] $} ,  
\label{conditional_equation_general}
\end{equation}
\begin{equation} 
\mathcal{L}_{L_{1}}\left ( G\right ) = \mathbb{E}_{Im_{j},Im_{i},z}\left [ \left \| Im_{i}-G\left ( Im_{j},z \right ) \right \|_{1} \right ] .
\label{Gen_loss_equation_general}
\end{equation}

\begin{figure}
     \centering
     \begin{subfigure}[b]{0.2\textwidth}
         \centering
         \includegraphics[width=\textwidth]{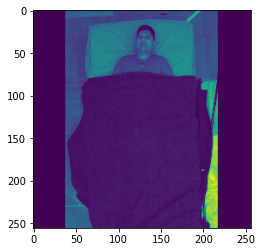}
         \caption{Visible}
         \label{fig:visible_modality}
     \end{subfigure}
     \hfill
     \begin{subfigure}[b]{0.2\textwidth}
         \centering
         \includegraphics[width=\textwidth]{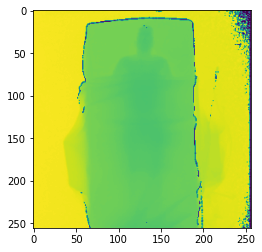}
         \caption{Depth}
         \label{fig:depth_modality}
     \end{subfigure}
     \hfill
     \begin{subfigure}[b]{0.2\textwidth}
         \centering
         \includegraphics[width=\textwidth]{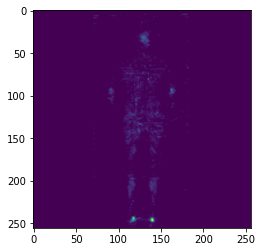}
         \caption{PM}
         \label{fig:PM_modality}
     \end{subfigure}
     \hfill
     \begin{subfigure}[b]{0.2\textwidth}
         \centering
         \includegraphics[width=\textwidth]{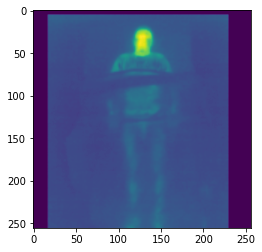}
         \caption{LWIR}
         \label{fig:LWIR_modality}
     \end{subfigure}
        \caption{Sample images from different modalities in the ``Danalab'' portion of the SLP dataset. Original image size: Visible (576 $\times$ 1024), Depth (424 $\times$ 512), PM (84 $\times$ 192), LWIR (120 $\times$ 160).
        For visualization purposes, images are resized to 256 $\times$ 256.}
        \label{ModalitySize_danalab}
\vspace{-4pt}
\end{figure}

\section{Experimental setup}
\label{experiments}

\subsection{Dataset}
\label{Dataset}
\begin{figure}
     \centering
     \begin{subfigure}[b]{0.3\textwidth}
         \centering
         \includegraphics[width=\textwidth]{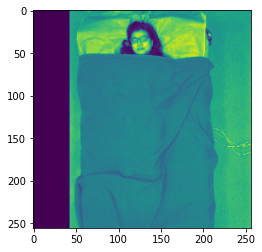}
         \caption{Visible}
         \label{fig:visible_modality}
     \end{subfigure}
     \hfill
     \begin{subfigure}[b]{0.3\textwidth}
         \centering
         \includegraphics[width=\textwidth]{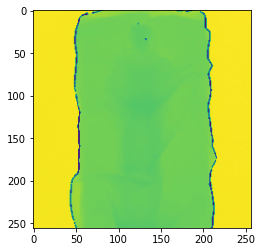}
         \caption{Depth}
         \label{fig:depth_modality}
     \end{subfigure}
     \hfill
     \begin{subfigure}[b]{0.3\textwidth}
         \centering
         \includegraphics[width=\textwidth]{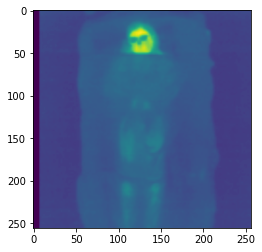}
         \caption{LWIR}
         \label{fig:LWIR_modality}
     \end{subfigure}
        \caption{Sample images from different modalities in the ``SimLab'' portion of the SLP dataset. Original image size: Visible (896 $\times$ 1600), Depth (424 $\times$ 512), LWIR (120 $\times$ 160).
        For visualization purposes, images are resized to 256 $\times$ 256.}
        \label{ModalitySize_simlab}
\vspace{-4pt}
\end{figure}

We used the publicly available SLP dataset~\cite{liu20120simultaneously,liu2019seeing} to train and test the fusion models outlined in Section \ref{fusion_appropach},
and assess these models' ability to generalize.
This dataset has patient pose images taken simultaneously from four different modalities: 1) Visible, 2) LWIR, 3) Depth and 4) Pressure mattress. In addition, data is captured under three different cover conditions: 1) uncovered, 2) bed sheet (covered 1), and 3) blanket (covered 2). 
This dataset consists of 2 portions, ``DanaLab" (Fig.~\ref{ModalitySize_danalab}) and ``SimLab" (Fig.~\ref{ModalitySize_simlab}). Images for ``DanaLab" has been collected in a home setting, whereas images for ``SimLab" has been collected in a hospital setting. Compared to ``DanaLab", images for ``SimLab" have different ceiling heights, different beds (commercial Hill-Rom hospital beds), different types of blankets (various brands and colors) and different participants \cite{liu20120simultaneously}. ``DanaLab" is used during the training and validation portions of model development, and the model is tested on the ``SimLab" set to evaluate its generalizability.

The dataset contains ground truth annotation for $(x,y)$ coordinates of 14 human body joints. These 14 joints are: left and right ankles, left and right knees, left and right hips, left and right wrists, left and right elbows, left and right shoulders, thorax, and the head.
Different combinations of modalities available in the SLP dataset were used to train the proposed feature fusion models. 
A summary of the dataset is presented in Table \ref{table:1}.

\vspace{-5pt}
\subsection{Evaluation protocol}
We benchmark our method using the percentage of correct keypoints at 0.5 (PCKh@0.5). In PCKh@0.5, an estimated joint of a subject is considered correct if the distance between the ground truth and the predicted joint is less than 50\% of the head bone length of the subject. Head bone length is the distance between the keypoints associated with head and thorax. The percentage of such correct estimations is calculated per joint~\cite{liu20120simultaneously}.

We also use 2D joint visualizations for qualitative validation of the pose estimations of the model. The effectiveness of the proposed models is validated on both the ``SimLab" and ``DanaLab" portions of the SLP dataset using the PCKh@0.5 metric.

\begin{table}[!t]
\caption{Summary of SLP dataset specifications.}
\vspace{1pt}
\centering
\resizebox{0.85\textwidth}{!}{%
\label{table:1}
\begin{tabular}{l l c}
\toprule
Dataset & Available Modalities & Number of Subjects \\
\midrule
SLP (DanaLab portion) & Visible, LWIR, Depth, Pressure Image & 102 adults\\
SLP (SimLab portion) & Visible, LWIR, Depth & 7 adults\\
\bottomrule
\end{tabular}}
\end{table}

\vspace{-5pt}
\subsection{Preprocessing and training for fusion strategies of multimodal images}

As described in Fig.~\ref{ModalitySize_danalab}, visible, depth, LWIR, and pressure images in the SLP dataset have different resolutions. Hence, images are resized to 256 $\times$ 256, before being fed into the model. 
Each fusion subset that includes depth images uses square bounding boxes to mitigate the severe noise that is present in the corners of the background of the depth images. Multimodal images are aligned with the modality with the highest available resolution. This is done to avoid any potential information loss caused by transformations for alignment. All images are subjected to normalization using Equation \ref{normalization_function}, 
\begin{equation} 
output_{c} = (input _{c} - mean_{c}) /std_{c} . 
\label{normalization_function}
\end{equation}
where $mean_{c}$ is the per-channel mean and $std_{c}$ is the per-channel standard deviation. This is done to avoid undesirable effects from uneven illumination and object surface properties, such as the color and texture of the blankets and clothes.

The HRNet models for single modalities and all proposed fusion models were trained on the SLP dataset for 100 epochs with a batch size of 64. 
The first 90 subjects of the SLP dataset are used as the training and validation set, while the following 12 subjects are used as the test set, similar to \cite{liu20120simultaneously}. The weights are initialized using normal distributions for all experiments conducted with the SLP dataset.

\vspace{-5pt}
\subsection{Preprocessing and training for the reconstruction of visible images}

Visible images are cropped to center the subject before training the pix2pix model \cite{pix2pix2017}. Then each LWIR image is aligned to the corresponding cropped visible image and finally, both images are resized to 100 $\times$ 256. This results in LWIR images being enlarged and the visible images being downsampled. 
Consequently, the resulting synthetic visible images are smaller than the those used to train the pose estimation model, which was used to evaluate the quality of synthetic visible images. All resizing operations, including resizing due to modality alignment, use bi-linear interpolation.
Since the cGAN requires square images, the previously aligned images are resized to 256 $\times$ 256 without cropping before being fed to the model. 

The cGAN model is trained and tested by pairing LWIR images of all cover conditions with the relevant uncovered visible images. The model is trained for 200 epochs with a constant learning rate of 0.0002 for the first 100 epochs. The learning rate was then set to linearly decay for the second 100 epochs, such that the learning rate approaches zero when the last epoch is reached.

Each generated cropped visible image is placed on a white background using the bounding box coordinates. These bounding box coordinates have been derived using the joint annotations. The bounding box coordinates are incorporated to accurately localize the human body within the input image. Then these processed visible images will be paired with respective LWIR images and fed to the fusion model for the pose estimation.

\vspace{-5pt}

\section{Results and discussion}

\subsection{Uni-modal baseline}
We first trained HRNet on the SLP dataset using each individual modality to reproduce the results recorded in the state-of-the-art \cite{liu20120simultaneously}.
The obtained results are recorded in Table~\ref{table:2}. 
The column ``SoTA'' contains the results reported in~\cite{liu20120simultaneously}. Note that~\cite{liu20120simultaneously} did not include results for uni-modal training for the visible modality.

\begin{table}[!t]
\caption{PCKh@0.5 values obtained when HRNet is trained on individual modes of the SLP dataset for 100 epochs (transfer learning).}
\vspace{1pt}
\centering
\resizebox{0.34\textwidth}{!}{%
\label{table:2}
\begin{tabular}{l c c}
\toprule
Modalities & Ours & SoTA \cite{liu20120simultaneously}\\
\midrule
LWIR & 93.2 & 93.4\\
Depth & 96.1 & 96.4\\
PM & 90.3 & 84.3\\
Visible & 93.7 & -\\
\bottomrule
\end{tabular}}
\vspace{-4pt}
\end{table}

\begin{table}[!t]
\caption{PCKh@0.5 values obtained when multiple modalities are fused using the proposed techniques of addition and concatenation.}
\vspace{1pt}
\centering
\resizebox{0.85\textwidth}{!}{%
\label{table:3}
\begin{tabular}{l c c c c c c c c}
\toprule
\multirow{3}{*}{Fusion stage} & \multirow{3}{*}{Fusion type} & Visible & Visible & Visible & Depth & Depth & LWIR & \\
 & & and & and & and & and & and & and & Average \\ 
 & & LWIR & Depth & PM & LWIR & PM & PM & \\
\midrule
3 & Concatenation & 94.9 & 94.2 & 91.9 & 93.8 & 94.9 & 83.8 & \textbf{92.25} \\
3 & Addition & 94.6 & 93.8 & 91.5 & 93.6 & 94.9 & 84.2 & \textbf{92.10}\\
2 & Concatenation & 94.4 & 95.5 & 90.5 & 93.3 & 94.1 & 81.5 & \textbf{91.55}\\
2 & Addition & 93.4 & 94.6 & 91.1 & 93.3 & 94.0 & 81.6 & \textbf{91.33}\\
\bottomrule
\end{tabular}}
\vspace{-4pt}
\end{table}

\subsection{Fusion: addition and concatenation}
While the presented model can be extended to support more modalities in combination, in this paper, we aim for an effective fusion of two modalities such that the models can be utilized in a low resource clinical environment.
Table~\ref{table:3} displays the PCKh@0.5 values obtained for different combinations of modalities using the fusion techniques introduced in Subsection~\ref{add_desc} and Subsection~\ref{concat_desc}. 

To examine the effect of the granularity of the features extracted on fusion performance, we selected stage 2 and stage 3 of HRNet~\cite{8953615}. 
It is observed that not all techniques outperform the results obtained for individual modalities. However, each bi-modal fusion model outperforms the poorest performing constituent modality except for LWIR-PM (fusion of modalities with the smallest dimensions). In particular, we observe that when depth is included, the fused system cannot match the performance of the depth mode alone. This shows that the modality with the lowest performance in each combination has reaped the benefit of the features extracted from the other modality to improve its performance.

From Table \ref{table:3}, it is also evident that for both stages $2$ and $3$, the concatenation technique has performed better in most cases, benefiting from the improved preservation of localized semantic information and the trained 1x1 convolutional layer. Furthermore, during the addition process, the addition of potentially misleading features from one modality with informative features with the other modality causes a reduction in the overall performance of the fusion model. In the concatenation technique, the learned convolutional layer that combines the concatenated features before the next stage of HRNet avoids this issue.

\begin{table}[!t]
\caption{PCKh@0.5 values obtained when multiple modalities are fused using the proposed \textit{weighted fusion} technique.}
\vspace{1pt}
\centering
\resizebox{0.85\textwidth}{!}{%
\label{table:4}
\begin{tabular}{l c c c c c c c c}
\toprule
\multirow{3}{*}{Fusion stage} & \multirow{3}{*}{Fusion type} & Visible & Visible & Visible & Depth & Depth & LWIR & \\
 & & and & and & and & and & and & and & Average\\ 
 & & LWIR & Depth & PM & LWIR & PM & PM & \\
\midrule
3 & Concatenation & 95.1 & 94.8 & 92.4 & 94.4 & 95.6 & 85.6 & \textbf{92.98} \\
3 & Addition & 95.0 & 94.6 & 92.5 & 95.0 & 95.8 & 85.3 & \textbf{93.03}\\
2 & Concatenation & 95.1 & 95.6 & 91.2 & 94.5 & 95.3 & 93.4 & \textbf{94.35}\\
2 & Addition & 95.2 & 95.8 & 89.8 & 94.1 & 95.3 & 83.7 & \textbf{92.32}\\
\bottomrule
\end{tabular}}
\vspace{-4pt}
\end{table}

\subsection{Fusion: learned modal weights}
As described in Subsection~\ref{weighted_desc}, the usefulness of the features extracted from different modalities for the final prediction is not equal. 
To better exploit the strengths of individual modes, we consider a ``weighted" fusion technique to allow the model to learn which feature maps should be positively weighted and which feature maps should be negatively weighted to minimize the loss. Results of the fusion experiments with trainable weights are available in Table~\ref{table:4}. When compared with Table~\ref{table:3}, it can be observed that the proposed weighted fusion technique yields better results with both addition and concatenation except for Visible-PM-Stage 2 Addition. It should be noted that during the training of these models, only the weights of the layers following stage $N$ (Equation \ref{weighted_add} and Equation \ref{weighted_concat}) are updated.

\subsection{Fusion: end-to-end learning}
The results related to end-to-end training of the fusion model are reported in Table \ref{table:5}. The end-to-end training of the fusion model has produced better results compared to the state-of-the-art uni-modal models, and the other fusion models presented in Tables \ref{table:3} and \ref{table:4}. The freedom given to the model to adjust weights before the feature extraction has resulted in a better fusion of features from both modalities.
Also, usage of spatial dropout has helped to avoid overfitting, and promote generalisation. But the end-to-end training of the fusion model is computationally intensive compared to earlier approaches, where only the weights of the layers following stage $N$ have been updated. The training time of the end-to-end fusion model is up to twice that of the earlier approaches. However, depending on the available computational resources and the required accuracy level of the specific application, the  number of layers that are subjected for the training can be easily varied in this model.  


\begin{table}[!t]
\caption{PCKh@0.5 values obtained when multiple modalities are fused using the proposed \textit{end-to-end fusion} technique.}
\vspace{1pt}
\centering
\resizebox{0.85\textwidth}{!}{%
\label{table:5}
\begin{tabular}{l c c c c c c c c}
\toprule
\multirow{3}{*}{Fusion stage} & \multirow{3}{*}{Fusion type} & Visible & Visible & Visible & Depth & Depth & LWIR & \\
 & & and & and & and & and & and & and & Average\\ 
 & & LWIR & Depth & PM & LWIR & PM & PM & \\
\midrule
3 & Concatenation & 96.0 & 96.9 & 97.1 & 97.0 & 97.5 & 96.1 & \textbf{96.77} \\
3 & Addition & 95.9 & 96.9 & 97.2 & 97.1 & 97.5 & 96.0 & \textbf{96.77}\\
2 & Concatenation & 95.8 & 97.0 & 97.2 & 97.0 & 97.6 & 95.9 & \textbf{96.75}\\
2 & Addition & 96.0 & 96.8 & 96.9 & 97.2 & 97.2 & 95.7 & \textbf{96.63}\\
\bottomrule
\end{tabular}}
\vspace{-4pt}
\end{table}


\begin{table}[!t]
\caption{Results of the validation done using the ``SimLab" portion of the SLP dataset.} 
\vspace{1pt}
\centering
\resizebox{0.55\textwidth}{!}{%
\label{table:7}
\begin{tabular}{l c}
\toprule
Modalities & PCKh@0.5 \\
\midrule
Depth & 95.3 \\
LWIR & 95.1 \\
Visible & 87.8 \\
\midrule
Visible-depth 3 Concatenation (end-to-end) & 96.7 \\
Depth-LWIR 3 Concatenation (end-to-end) & 97.3\\
Visible-LWIR 3 Concatenation (end-to-end) & 95.1 \\
\bottomrule
\end{tabular}}
\vspace{-6pt}
\end{table}

\begin{figure}[!t]
\centering
\includegraphics[width=0.9\linewidth]{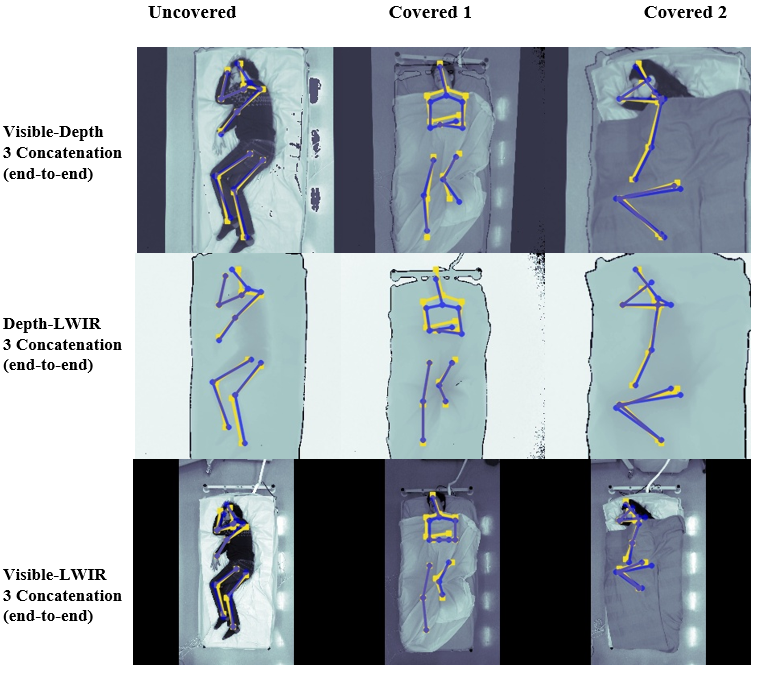}
\vspace{-3pt}
\caption{Qualitative results of in-bed pose estimation for Fusion models on samples from the “SimLab” portion of the SLP dataset. The yellow skeletons depict the ground truth and the blue skeletons depict the prediction.}
\label{simLab_results}
\vspace{-12pt}
\end{figure}

\subsection{Fusion: generalization}
To evaluate the generalization ability of the proposed methods, models were trained on ``DanaLab'' and were tested on the ``SimLab'' portions of the SLP dataset. Images in ``SimLab" are different from those in ``DanaLab", since the setting of the images is a simulated hospital environment, which contains changes in illumination and other factors discussed in Section \ref{Dataset}, that are not observed during training. Hence, validation tests were executed on ``SimLab" samples to prove the generalizability of the proposed models. The results corresponding to these experiments are recorded in Table \ref{table:7}. The best performing setting (\textit{i.e.}, the combination of the stage and the fusion technique) from previous experiments was used here for testing. 

From Table~\ref{table:7} it is evident that the end-to-end learning fusion model proposed in this paper has obtained equal or better performance than uni-modal approaches. Qualitative results for this evaluation are presented in Fig.~\ref{simLab_results}.

\begin{figure}[!t]
\centering
\includegraphics[width=0.5\linewidth]{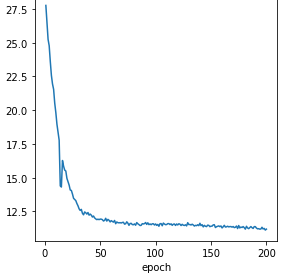}
\vspace{-3pt}
\caption{LWIR to visible generator L1 loss curve.}
\label{generator_l1_loss}
\vspace{-6pt}
\end{figure}

\begin{figure*}[!t]
\centering
\includegraphics[width=0.9\linewidth]{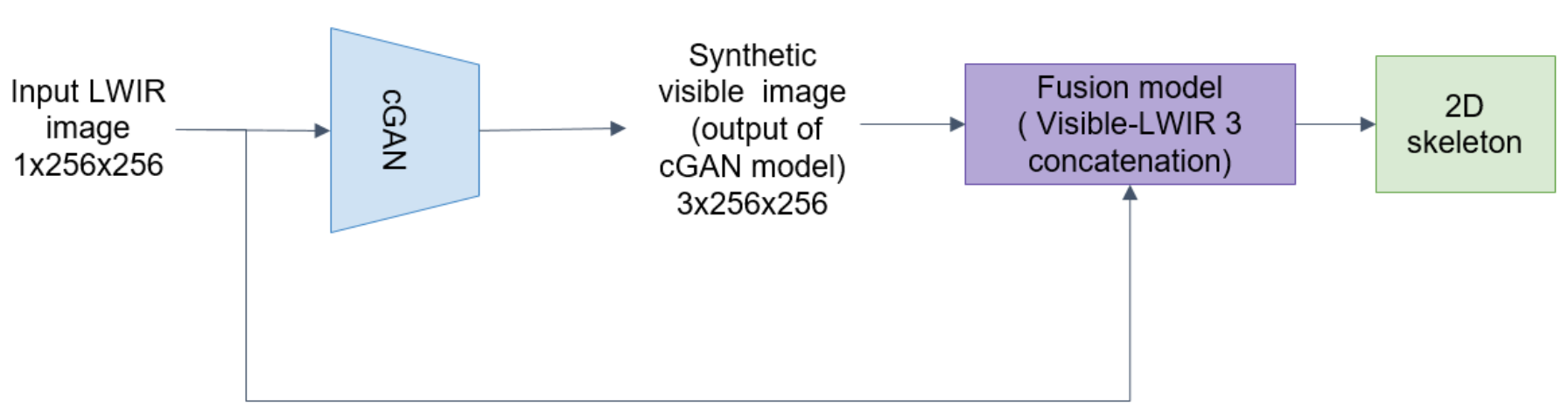}
\vspace{-2pt}
  \caption{The pipeline of using synthesized visble images.  Input LWIR images are fed to the pre-trained cGAN which generates the synthesized visible images. Then these synthesized visible images are fed to the fusion model along with the corresponding LWIR images for the pose estimation.}
\label{reconstruction-pipeline}
\vspace{-8pt}
\end{figure*}

\subsection{Fusion: synthesized visible images}

The $L_{1}$ error between
generated visible images and real visible images associated with cGAN (Equation \ref{Gen_loss_equation} and Equation \ref{objective_function}) provides a quantitative measurement of the cGAN's ability to reconstruct the visible images from LWIR images. Fig.~\ref{generator_l1_loss} depicts this reconstruction error during model training, where it can be seen the loss has gradually decreased as the model obtains the ability to better reconstruct the visible images.

The LWIR images for this evaluation were selected so that the pose estimation model had not previously seen their paired visible image. Those images were used to generate synthetic visible images. Then the synthetic images were processed and paired up with respective LWIR images and fed into a pre-trained Visible-LWIR stage 3 concatenation fusion pose estimation model, trained end-to-end for pose prediction (See Fig.~\ref{reconstruction-pipeline}).

\begin{figure}[!t]
\centering

\includegraphics[width=0.95\linewidth]{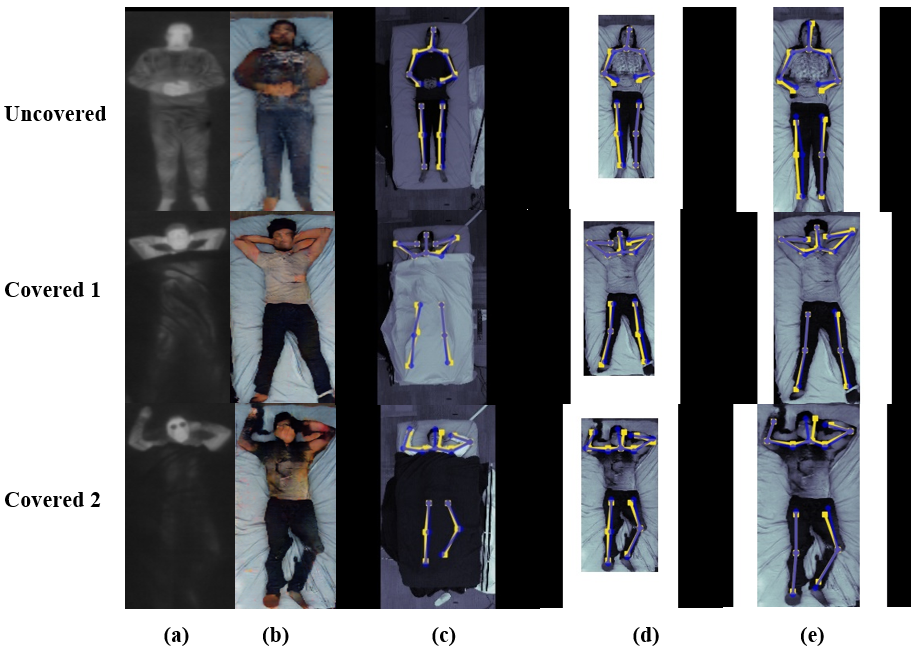}
\vspace{-3pt}
\caption{
Qualitative results showing visible images generated by cGAN model and the pose estimations for those images (using a Visible-LWIR stage 3 concatenation fusion model (end-to-end learning)). 
(a) Original LWIR image, (b) Generated visible image, (c) Prediction for original visible image, (d) Prediction for generated visible image cropped without the square bounding box, (e) Prediction for generated visible image cropped with a square bounding box. The yellow skeletons depict the ground truth and the blue skeletons depict the prediction.}
\label{cGAN_results}
\vspace{-10pt}
\end{figure}

The performance was evaluated using the PCKh@0.5 metric. This model was selected because it yielded the best results for the Visible-LWIR modalities when trained with the original data (\textit{i.e.}, based on Table \ref{table:3}, Table \ref{table:4} and Table \ref{table:5}). 
When the synthetic visible images were fed to the pose estimation model, the effect of cropping the image using square bounding boxes was also tested. The results of these experiments are shown in Table \ref{table:8}. Some qualitative results are shown in Fig.~\ref{cGAN_results}.

\begin{table}[!t]
\caption{Comparison between PCKh@0.5 values per joint obtained by the best performing models on real visible and LWIR modality combination of each technique and that of obtained by running pose estimation tests on a Visible-LWIR stage 3 concatenation (end-to-end learning) model (the best performing model out of all techniques) using synthetic visible images generated from LWIR and the corresponding LWIR image. Note that the results obtained using synthetic data are obtained using a model that has been trained on real data only.}
\vspace{1pt}
\centering
\resizebox{1\textwidth}{!}{%
\label{table:8}
\begin{tabular}{l c c c c c c }
\toprule
\multirow{2}{*}{Joints} & \multicolumn{3}{c}{Fusion with} & & \multicolumn{2}{c}{Fusion with} \\
& \multicolumn{3}{c}{real visible and LWIR images} & & \multicolumn{2}{c}{synthetic visible and LWIR images} \\
\cmidrule{2-4}
\cmidrule{6-7}
& addition and & learned model & end-to-end & & cropping based on & without cropping\\
& concatenation & weights & learning & & square BB & based on square BB\\
\midrule
Right Ankle & 96.0 & 96.8 & 97.4 & & 89.3 & 93.5 \\
Right Knee & 96.5 & 97.5 & 98.0 & & 89.0 & 93.5 \\
Right Hip & 93.0 & 93.0 & 93.3 & & 82.1 & 87.9 \\
Left Hip & 93.1 & 92.5 & 93.8 & & 82.3. & 87.8 \\
Left Knee & 96.7 & 96.8 & 97.0 & & 88.5 & 93.2 \\
Left Ankle & 96.5 & 97.0 & 98.0 & & 88.5 & 92.7 \\
Right Wrist & 89.6 & 90.2 & 92.9 & & 76.1 & 81.4 \\
Right Elbow & 93.3 & 92.4 & 95.4 & & 80.7 & 87.9 \\
Right Shoulder & 94.1 & 95.1 & 94.8 & & 89.0 & 90.9 \\
Left Shoulder & 96.1 & 95.6 & 95.6 & & 88.9 & 90.6 \\
Left Elbow & 94.4 & 95.0 & 96.8 & & 81.8 & 89.2 \\
Left Wrist & 90.6 & 90.7 & 93.6 & & 74.5 & 79.1 \\
Thorax & 99.3 & 99.3 & 98.8 & & 98.3 & 98.1 \\
Head & 98.7 & 98.8 & 99.1 & & 96.9 & 97.5 \\
Total & \textbf{94.9} & \textbf{95.1} & \textbf{96.0} & & \textbf{86.1} & \textbf{90.2}\\
\bottomrule
\end{tabular}}
\vspace{-4pt}
\end{table}

Results obtained by running pose estimation without using bounding boxes are better than those obtained when using square bounding boxes. A square with the subject at the center is considered when a square bounding box is used. The Visible-LWIR-Stage-3-Concatenation (end-to-end learning) fusion model used for the pose estimation testing was also trained using unbounded images. This is why pose estimation testing without using bounding boxes yields better results compared to testing using square bounding boxes. 

While the results obtained with the synthesized visible images do not outperform the results obtained with real visible images, from the qualitative and quantitative results it can be seen that the synthesized visible images have enabled pose estimation with an average of 90.2\% PCKh@0.5. The main problem of in-bed pose estimation is the high probability of all or part of the subject being covered by a sheet or blanket while on the bed (see Fig.~\ref{cGAN_results}). Some sensing modalities may provide other forms of information for pose inference, however colour information is typically not available from such modes and visible images are rich with texture compared to the images from other modalities. This work is a pilot study for pose estimation under information loss. Thus, a comparable performance for learning cross-modality representations can be obtained when deploying AI-based methods with our methodology. Moreover, in traditional applications for sleep monitoring~\cite{mohammadi2020transfer,mohammadi2018sleep}  and in-bed fall prevention~\cite{kramer2020automated,wang2021video} the classification is conducted on the global pose, and it is not required to have the precise keypoint locations. 

It should also be noted that the fusion model has not been trained on synthesized images, as the objective of this experiment is to evaluate the performance of the existing pre-trained model when the real visible images are not available in the deployed system.

Considering our prior research and other previous work, no existing method has used synthesized visible images in multi-modal pose estimation. The approach proposed in this work makes fusion models suitable for scenarios where visible images are not available in the dataset, and also enables privacy-preserving pose estimation. 

\begin{figure}[!t]
\centering
\includegraphics[width=.32\textwidth]{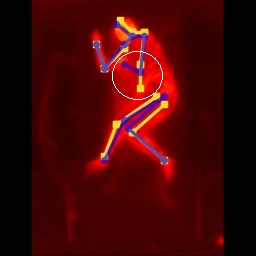}
\includegraphics[width=.32\textwidth]{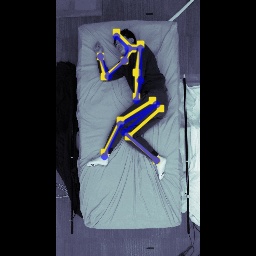}
\includegraphics[width=.32\textwidth]{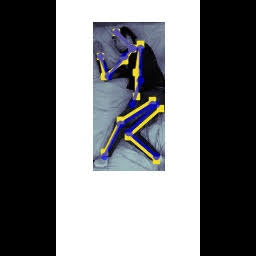}
\vspace{-2pt}
\caption{Pose ambiguities in LWIR images caused by heat residuals. The first column depicts the LWIR image, the second column depicts the corresponding visible image, and the third column depicts the synthesized visible image. The yellow skeletons depict the ground truth and the blue skeletons depict the prediction. The white circle depicts the area affected by the residual heat. }
\label{residual_lwir}
\vspace{-10pt}
\end{figure}

Residual heat in LWIR images can mislead the pose estimation algorithms, whereas the visible images are unaffected by this issue. Fig.~\ref{residual_lwir} shows examples where residual heat misleads the pose detection, and in such scenarios incorporating knowledge from visible images can favourably affect the outcome. In Fig.~\ref{residual_lwir} the residual heat has adversely affect  the estimation of the left wrist. 
When the normalized $L_{2}$ error of the left wrist is considered, it is 1.56 when the LWIR image is used where as it is 0.13 on the real visible image and 0.39 on the synthesized visible image.
Furthermore, the average normalized $L_{2}$ error for the LWIR modality is 0.36 and it is  0.18 for both real visible images and the synthesized visible images. 
From the quantitative and qualitative results, it can be observed that the end-to-end pipeline, which contains two main components (visible image reconstruction and fusion-based pose estimation), can robustly perform in-bed pose estimation while preserving the privacy of the subjects.

\section{Conclusion}
\label{conclusion}

Research into in-bed human pose estimation has grown due to its applicability to many healthcare applications. In this paper, we explored the effective use of multi-modal data to enable robust in-bed pose estimation. To that end, several intermediate feature fusion approaches with the HRNet model as the backbone were proposed. We have introduced Addition, Concatenation, Fusion via learned modal weights, and an end-to-end fully trainable approach to analyse the effect of feature fusion for in-bed human pose estimation. We demonstrated that our end-to-end feature fusion techniques outperform the performance of single modalities. We demonstrated the generalization ability of the fusion models through evaluations between two different data settings.
We also demonstrated the use of reconstructed visible images from existing LWIR images for in-bed human pose estimation using the best performing multi-modal fusion technique. This will be beneficial for scenarios where access to cameras to capture visible images is restricted. This will also eliminate privacy violation concerns that come with capturing raw visible images in a hospital environment. Thus, the applications of in-bed pose estimation in hospital environments can be widened by utilizing the frameworks proposed in this paper.
Possible future improvements on the current work include using low-resolution pose estimation techniques to reduce the ill-effects caused by low sensor resolutions and differences in sensor resolution, and using more than two modalities for fusion.

\bibliographystyle{splncs04}
\bibliography{egbib}
\end{document}